\documentclass[conference]{IEEEtran}
\IEEEoverridecommandlockouts
\usepackage{cite}
\usepackage{amsmath,amssymb,amsfonts}
\usepackage{algorithmic}
\usepackage{graphicx}
\usepackage{textcomp}
\def\BibTeX{{\rm B\kern-.05em{\sc i\kern-.025em b}\kern-.08em
    T\kern-.1667em\lower.7ex\hbox{E}\kern-.125emX}}

\usepackage{booktabs}
\usepackage{multirow}
\usepackage{bbding}
\usepackage[table]{xcolor}
\usepackage{epstopdf}

\begin{document}

\title{ICFRNet: Image Complexity Prior Guided Feature Refinement for Real-time Semantic Segmentation\\
\thanks{Corresponding author: Jinglei Shi.}
\thanks{This work was supported by the Natural Science Foundation of Tianjin, China (NO. 20JCJQJC00020), the National Natural Science Foundation of China (Nos. 62302240), Fundamental Research Funds for the Central Universities, Supercomputing Center of Nankai University (NKSC), the Bulgarian Ministry of Education and Science under the National Research Programme “Young scientists and postdoctoral students - 2” approved by DCM 206 / 07.04.2022.}
}

%

\author{
\IEEEauthorblockN{\textit{Xin Zhang}\textsuperscript{1,2}, \textit{Teodor Boyadzhiev}\textsuperscript{3}, \textit{Jinglei Shi}$^{\ast}$\textsuperscript{1}, \textit{and Jufeng Yang}\textsuperscript{1,2}}
\\
\IEEEauthorblockA{\textsuperscript{1} VCIP \& TMCC \& DISSec, College of Computer Science, Nankai University, Tianjin, China}
\IEEEauthorblockA{\textsuperscript{2} Nankai International Advanced Research Institute (SHENZHEN· FUTIAN), Shenzhen, China}
\IEEEauthorblockA{\textsuperscript{3} Institute of Mathematics and Informatics, Bulgarian Academy of Sciences, Sofia, Bulgaria}
\IEEEauthorblockA{Email: zhangxin.nk@mail.nankai.edu.cn, t.boyadzhiev@math.bas.bg, \{jinglei.shi, yangjufeng\}@nankai.edu.cn}}



\maketitle

\begin{abstract}
In this paper, we leverage image complexity as a prior for refining segmentation features to achieve accurate real-time semantic segmentation.
The design philosophy is based on the observation that different pixel regions within an image exhibit varying levels of complexity, with higher complexities posing a greater challenge for accurate segmentation. 
We thus introduce image complexity as prior guidance and propose the Image Complexity prior-guided Feature Refinement Network (ICFRNet). 
This network aggregates both complexity and segmentation features to produce an attention map for refining segmentation features within an Image Complexity Guided Attention (ICGA) module.
We optimize the network in terms of both segmentation and image complexity prediction tasks with a combined loss function. 
Experimental results on the Cityscapes and CamViD datasets have shown that our ICFRNet achieves higher accuracy with a competitive efficiency for real-time segmentation. 
\end{abstract}


\section{Introduction}
\label{sec:intro}

Real-time semantic segmentation stands as an essential task within the field of computer vision~\cite{9102769,9428381}. 
In order to fulfill the requirements of real-time calculations, some methods~\cite{zhang2021efrnet,chen2023icanet,yu2021bisenet,tsai2023bisenet} hinge on the implementation of lightweight network architectures, and finally attain speeds approaching 100 \textit{fps} or higher on a single GPU.
However, these methods often compromise segmentation accuracy and struggle to handle subtle or complex structures within the scene.
\begin{figure}[t]
    \centering
    \includegraphics*[width=0.48\textwidth]{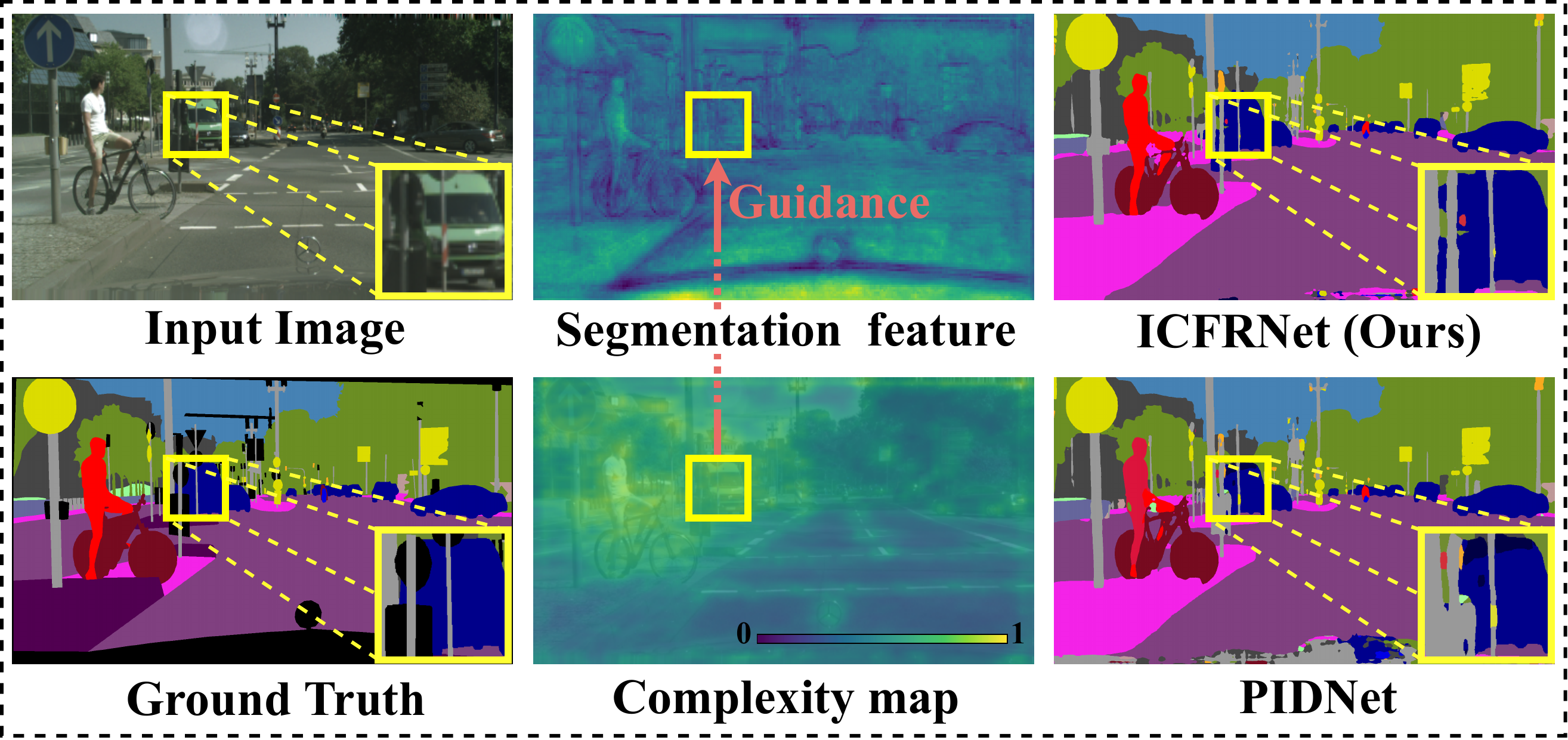}
    \caption{Presentation of the relationship between image complexity and semantic segmentation.
    The regions of high complexity marked with yellow boxes contain more details and the pixels are prone to misclassification.
    With the image complexity guidance, our ICFRNet pays more attention to the complex pixel regions.
    }
    \vspace{-0.4cm}
    \label{fig:motivation}
\end{figure}
Our objective is to enhance segmentation accuracy, particularly for challenging regions, but without significantly lowering efficiency for real-time processing.

Given the fact that semantic segmentation is essentially to assign a class label to each pixel, the final accuracy is profoundly decided by the performance of this classification process.
Previous methods~\cite{paszke2016enet, Romera2018ERFNetER, li2019dfanet} have concentrated on utilizing efficient convolution operations to diminish computational complexity, and employ sophisticated features aggregation mechanisms to benefit efficient segmentation.
While others designed the two-branch pipeline~\cite{yu2018bisenet} to capture spatial details and semantic information~\cite{yu2021bisenet, pan2022deep, xu2023pidnet} respectively for better segmentation performance. 
Despite the capability of these methods to produce decent segments by exploring spatial details and semantic information, they neglect the varying levels of classification difficulty of each pixel, resulting in limited performance when segmenting complex regions, as shown in Fig.~\ref{fig:motivation}.
And as explained in~\cite{feng2022ic9600}, such classification difficulty (or complexity) 
is primarily attributed to the increased information density inherent in these complex regions. 
It serves as a useful clue for pixel classification and has been proven helpful for the segmentation task.

However, unlike in~\cite{feng2022ic9600}, where the image complexity map only participates in the final supervision in the segmentation task, we take a further step by integrating image complexity as prior guidance for assigning varying levels of attention to the intermediate segmentation features.
\begin{figure*}[ht]
    \centering
    \includegraphics*[width=1\textwidth]{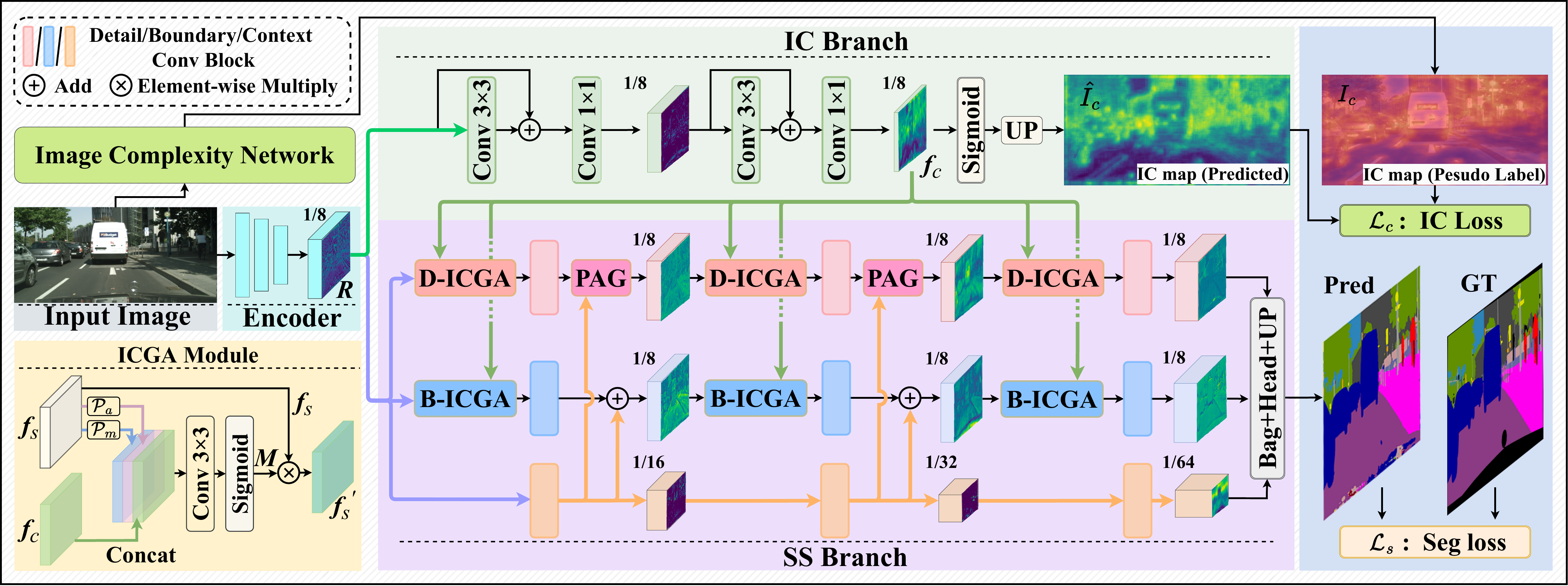}
    \caption{
    Illustration of the pipeline.
    The input image is processed through the encoder and the obtained latent features $R$ flow into two branches ($i.e.$, IC Branch and SS Branch).
    D- denotes the detail sub-branch and B- denotes the boundary sub-branch.
    The image complexity features are aggregated into the detail and boundary sub-branches for feature refinement through our ICGA module.
    Also, the semantic feature from the context sub-branch is fused into detail and boundary features through the PAG~\cite{xu2023pidnet} and adding operation~\cite{xu2023pidnet} separately.
    IC Loss $\mathcal{L}_{c}$ and Seg Loss $\mathcal{L}_{s}$ are the loss functions of IC Branch and SS Branch.
}
    \label{fig:pipeline}
\end{figure*}
Concretely, we construct a dual-task framework named ICFRNet that simultaneously predicts image complexity through the Image Complexity (IC) branch, and image segments through the Semantic Segmentation (SS) branch.
The IC branch is supervised with the pseudo labels generated by the Image Complexity Network (ICNet)~\cite{feng2022ic9600} via Mean Square Error (MSE) and Kullback-Leibler divergence (KL)-based loss function.
While the SS branch adopts PIDNet~\cite{xu2023pidnet} as basic structure, which includes three sub-branches (detail, context and boundary ones) and is supervised by Cross Entropy-based loss function.
To leverage complexity information for enhancing segmentation accuracy, we introduce an Image Complexity Guided Attention (ICGA) module within the SS branch. 
In this module, the intermediate image complexity features from the IC branch are incorporated into the SS branch. 
Firstly, these features are concatenated with pooled segmentation features and then aggregated through a convolution layer with $Sigmoid$ activation to generate a spatial attention map. 
\textcolor{black}{This attention map is subsequently applied to modulate the segmentation features ($i.e.$, detail features and boundary features), enabling the final prediction to account for semantic, spatial, boundary, and complexity information,} thus improving segmentation quality in especially complex regions. 
Experimental results have demonstrated that our method can effectively explore image complexity information to generate high-quality segments, even for subtle or complex structures that are challenging for other SOTA methods. 
Our contributions can be summarized as follows:
\begin{itemize}
    \item We propose an efficient way to apply image complexity information to the task of semantic segmentation. 
    Instead of simply using the complexity map to supervise the final prediction, we construct a dual-task framework that predicts both image complexity and segmentation.
    We further design a novel ICGA module within the SS branch that adopts the intermediate complexity features to guide the prediction of segments in a manner of spatial attention.
    \item We carried out extensive experiments on Cityscapes~\cite{cordts2016cityscapes} and CamViD~\cite{BrostowSFC:ECCV08} datasets to evaluate our method.
    The results demonstrate that our framework can effectively use image complexity to improve the segmentation accuracy, especially for subtle regions that are difficult to be correctly segmented, meanwhile maintaining low computational complexity for real-time inference.
\end{itemize}

\section{Related Work}

\subsection{Real-time Semantic Segmentation}

Achieving a trade-off between inference speed and accuracy is a key challenge in real-time semantic segmentation tasks.
Most of the existing methods have far exceeded the real-time demand. 
For instance, \cite{paszke2016enet,Romera2018ERFNetER,li2019dabnet} utilize dilated and factorized convolution to design a lightweight network for ensuring parameter compactness.
However, they compromise a lot on segmentation quality.
Later, BiSeNet~\cite{yu2018bisenet} proposed a two-branch pipeline and was developed by following works~\cite{yu2021bisenet,tsai2023bisenet,fan2021rethinking}, where one branch handles scene details and the other tackles the context information.
DDRNet~\cite{pan2022deep} further enhances the above feature representation by facilitating feature exchange between the detail branch and the context branch. 
Moreover, PIDNet~\cite{xu2023pidnet} extends this concept by introducing boundary features, resulting in a three-branch network structure comprising detail, context, and boundary branches. 
These works benefit from the two-branch structure and try to obtain improved segmentation results, while some of the pixel regions with high complexity still suffer from limited segmentation performance.

\subsection{Image Complexity}
The image complexity describes the amount of details and variety within an image.
Feng et al.~\cite{feng2022ic9600} provided a thorough exposition of image complexity and collected the first large-scale dataset (IC9600) on this subject.
They developed the Image Complexity Network (ICNet), a dual-stream architecture designed to evaluate overall image complexity as well as a pixel-level complexity distribution map. 
This image complexity map is initially applied to the loss function during training in~\cite{feng2022ic9600}.
However, the further potential of refining latent segmentation features directly with image complexity information is not explored. 

\begin{table*}[t]\footnotesize
    \renewcommand\arraystretch{1.3}
    \centering
    \begin{tabular}{p{3cm}<{\raggedright}p{1.8cm}<{\raggedright}p{1.5cm}<{\centering}p{1.5cm}<{\centering}p{1.5cm}<{\centering}p{1.8cm}<{\centering}p{1.4cm}<{\centering}p{1.4cm}<{\centering}}
        \toprule[1.5pt]
        \multirow{2}{*}{Methods} & \multirow{2}{*}{Resolution} & \multirow{2}{*}{GPU} & \multicolumn{2}{c}{mIoU~($\%$)$\uparrow$} & \multirow{2}{*}{FPS$\uparrow$} & \multirow{2}{*}{GFLOPs$\downarrow$} & \multirow{2}{*}{Params(M)$\downarrow$}\\
        \cline{4-5}
         & &&Val set & Test set & &&\\
        \midrule[1.2pt]
         ENet~\cite{paszke2016enet} & 640 $\times$ 360 & Titan X & - & 58.3 & 135.4 & 3.8 & 0.4\\
         MSFNet~\cite{si2019real} & 2048 $\times$ 1024 & RTX 2080Ti & - & 77.1(+18.8) & 41.0(--94.4)  & 96.8(+93.0) & -\\
         \hline
         DF2-Seg1~\cite{li2019partial} & 1536 $\times$ 768 & GTX 1080Ti & 75.9 & 74.8(--2.3) & 67.2(+26.2) & - & - \\
         DF2-Seg2~\cite{li2019partial} & 1536 $\times$ 768 & GTX 1080Ti & 76.9(+1.0) & 75.3(+0.5) & 56.3(--10.9) & - & - \\
         \hline
         SwiftNetRN-18~\cite{orsic2019defense} & 2048 $\times$ 1024 & GTX 1080Ti & 75.5(--1.4) & 75.4(+0.1) & 39.9(--16.4) & 104.0 & 11.8 \\
         SwiftNetRN-18~ens~\cite{orsic2019defense}  & 2048 $\times$ 1024 & GTX 1080Ti & - & 76.5(+1.1) & 18.4(--21.5) & 218(+114) & 24.7(+12.9) \\
         CABiNet~\cite{kumaar2021cabinet} & 2048 $\times$ 1024 & RTX 2080Ti & 76.6 & 75.9(--0.6) & 76.5(+58.1) & 12.0(--206) & 2.6(--22.1) \\
         \hline
         BiSeNetV2\dag\textsuperscript{*}~\cite{yu2021bisenet} & 1024 $\times$ 512 & GTX 1080Ti & 73.4 (--3.2) & 72.6 (--3.3) & 156.0 (+79.5) & 21.1 (+9.1) & 3.6((+1.0)\textsuperscript{\tiny$\triangle$}\\
         BiSeNetV2-L\dag\textsuperscript{*}~\cite{yu2021bisenet} & 1024 $\times$ 512 & GTX 1080Ti & 75.8(+2.4) & 75.3(+2.7) & 47.3(--108.7) & 118.5(+97.4) & -\\
         BiSeNetV3(STDC1)\dag\textsuperscript{*}~\cite{tsai2023bisenet} & 1024 $\times$ 512 & GTX 1080Ti & 73.4(--2.4) & 73.5(--1.8) & 244.3(+197) & - & -\\
         BiSeNetV3(STDC2)\dag\textsuperscript{*}~\cite{tsai2023bisenet} & 1024 $\times$ 512 & GTX 1080Ti & 74.6(+1.2) & 74.5(+1.0) & 180.3(--64.0) & - & -\\
         \hline
         STDC1-Seg75\textsuperscript{*}~\cite{fan2021rethinking} & 1536 $\times$ 768 & RTX 3090 & 74.5(--0.1) &  75.3(+0.8) & 74.8(--105.5) & 71.5\textsuperscript{\tiny$\triangle$} & 10.6\textsuperscript{\tiny$\triangle$} \\
         STDC2-Seg75\textsuperscript{*}~\cite{fan2021rethinking} & 1536 $\times$ 768 & RTX 3090 & 77.0(+2.5) &  76.8(+1.5) & 58.2(--16.6) & 131.2(+59.7)\textsuperscript{\tiny$\triangle$} & 18.7(+8.1)\textsuperscript{\tiny$\triangle$} \\
         
         PP-LiteSeg-T2~\cite{peng2022pp} & 1536 $\times$ 768 & RTX 3090 & 76.0(--1.0) &  74.9(--1.9) & 96.0(+37.8) & - & - \\
         \hline
         HyperSeg-M~\cite{nirkin2021hyperseg} & 1024 $\times$ 512 & RTX 3090 & 76.2(+0.2) & 75.8(+0.9) & 59.1(--36.9) & 7.5 & 10.1\\
         DDRNet-23-S~\cite{pan2022deep} & 2048 $\times$ 1024 & RTX 3090 & 77.8(+1.6) &  75.4(--0.4)\textsuperscript{\tiny$\triangle$} & 108.1(+49.0) & 36.3(+28.8) & 5.7(--4.4) \\
         PIDNet-S~\cite{xu2023pidnet} & 2048 $\times$ 1024  & RTX 3090 & 78.8(+1.0) &  76.9(+1.5)\textsuperscript{\tiny$\triangle$} & 93.2(--14.9) & 47.6(+11.3) & 7.6(+1.9) \\
         \hline
         \rowcolor{gray!20}\textbf{ICFRNet (Ours)} & 2048 $\times$ 1024  & RTX 3090 & \textbf{79.2}(+0.4) & \textbf{77.5}(+0.6) & 83.8(-9.4) & 55.8(+8.2) & 7.9(+0.3) \\
         \bottomrule[1.5pt]

    \end{tabular}
    \vspace{0.1cm}
    \caption{Comparison with state-of-the-art methods on the Cityscapes dataset. 
    Methods marked with \textsuperscript{*} are trained with both train and val set for testing. 
    \dag~means the method used TensorRT for acceleration. 
    Results with \textsuperscript{\tiny$\triangle$} means our implementation, which is not reported or under different experiment settings in their papers.
    The values within parentheses denote the disparity between the current method and the one presented in the preceding line with respect to the corresponding metric.
    This indicates the trade-off between all of the metrics.
    $\downarrow(\uparrow)$ means the smaller the better (the larger the better).
    }
    \label{tab:city} 
\end{table*}

\section{Methodology}

\subsection{Network Overview}
The overall architecture of our proposed method is illustrated in Fig.~\ref{fig:pipeline}, and it can be divided into three functional parts: a lightweight encoder, an Image Complexity (IC) branch, and a Semantic Segmentation (SS) branch. 

We adopt the first three layers of the encoder in PIDNet~\cite{xu2023pidnet} as our lightweight feature extractor $\mathcal F$. 
Given an input image $I \in \mathbb{R}^{3 \times H \times W}$, the feature extractor will output a latent pixel representation $R = \mathcal F(I)$ of size $C \times \frac{H}{8} \times \frac{W}{8}$ for the later processing, where $C$ is the number of channels. 
It's noteworthy that the encoder used in PIDNet only explores information for the task of segmentation, our encoder is followed by the IC branch and SS branch, which means the latent pixel representation $R$ will contain information for both two tasks.

Our IC branch aims at predicting an image complexity map $\hat{I_c} \in \mathbb{R}^{1 \times H \times W}$ from $R$. As shown in Fig.~\ref{fig:pipeline}, it consists of two cascaded residual blocks, with each block including two convolution layers with $3\times3$ and $1\times1$ kernels, and generates a complexity feature map $f_{c}$ of size $1 \times \frac{H}{8} \times \frac{W}{8}$. Final complexity image $\hat{I_c}$ is obtained by passing $f_{c}$ through a $Sigmoid$ activation function and bilinear upsampling operation.

While our SS branch takes PIDNet~\cite{xu2023pidnet} as basic structure, it is composed of three sub-branches to respectively capture detail, context and boundary information for predicting an image of segmentation $\hat{I_s} \in \mathbb{R}^{1 \times H \times W}$. Based on this fundamental pipeline design, we further propose an Image Complexity Guided Attention (ICGA) mechanism (detailed in Sec.~\ref{sec:ICGA}), which allows the prediction of segmentation effectively taking scene complexity information into consideration.

\subsection{Image Complexity Guided Attention Module}
\label{sec:ICGA}
 As demonstrated in~\cite{feng2022ic9600}, image complexity reflects the distribution of information density, and reveals classification difficulties between different pixel regions, thus being useful for various computer vision tasks. 
 However, in \cite{feng2022ic9600}, complexity maps are merely engaged in the calculation of loss function during the training of a segmentation network, and complexity information is insufficiently explored. 
 We instead incorporate the complexity features of the IC branch $f_c$ into the SS branch and accordingly propose ICGA modules to more effectively make use of the complexity information.
 
 More precisely, like shown in Fig.~\ref{fig:pipeline}, 
 ICGA modules follow behind feature maps in both detail and boundary sub-branches ($i.e.$, D-ICGA / B-ICGA). 
 Each ICGA module takes the image complexity features $f_c$ and intermediate segmentation features $f_s$ ($i.e.$, detail features and boundary features) as input.
 Similar to~\cite{Woo_2018_ECCV}, the complexity features $f_{c}$ are first concatenated with the max- and average-pooled segmentation features, and then aggregated through a $3 \times 3$ convolution layer followed by $Sigmoid$ function to generate a spatial attention map $M$:
 \begin{equation}
     M = 
 \mathcal{H}([\mathcal{P}_{m}(f_{s}), ~\mathcal{P}_{a}(f_{s}), ~f_{c}]),
\end{equation}
where $\mathcal{H}$ is the $3 \times 3$ convolution layer followed by a $Sigmoid$ function, and $\mathcal{[\cdot,\cdot]}$ is the concatenation operation along channel dimension. 
$\mathcal{P}_{m}~and ~\mathcal{P}_{a}$ are global maximum pooling and global average pooling along channel dimension respectively.
The pooled segmentation features are engaged as auxiliary attention information and concatenated with image complexity features together to alleviate the over-modulation of segmentation features by the image complexity features.

This attention map is finally used to modulate the segmentation features $f_{s}$ to produce refined segmentation features $f'_{s}$ as:
 \begin{equation}
     f'_{s} = 
 M \times f_{s}.
\end{equation}
The refined features $f'_{s}$ will replace original segmentation features $f_{s}$ to participate in the following processing. 
Let us note that we only integrate the ICGA module into the detail and boundary sub-branches.
We exclude it from the context sub-branch as the semantic features concentrate more on consistent contextual information, while complexity features reveal more spatial complexity details. 
Moreover, our ICGA module could be easily applied to other segmentation networks or to networks for other computer vision tasks that involve complexity features.

\subsection{Dual-task Framework Optimization}

Our proposed framework is optimized in a dual-task manner with a combined loss function.
For the IC branch, the predicted image complexity map $\hat{I_c}$ is generated with $f_{c}$ as:
\begin{equation}
    \hat{I_c} = UP_{8\times}(\sigma (f_{c})),
\end{equation}
where $\sigma$ is the $Sigmoid$ function and $I_c$ is with size of $1 \times H \times W$.
We design the loss function of the image complexity prediction task as follows:
\begin{gather}
    \mathcal{L}_1 = MSE(\hat{I_c},~I_{c}), \\
    \mathcal{L}_2 = KL(\mathcal{LS}(\frac{\mathcal{M}(\hat{I_c})}{T}),~\mathcal{S}(\frac{\mathcal{M}(I_{c})}{T})),\\
    \mathcal{L}_{c} = \lambda_1  \mathcal{L}_1 + \lambda_2  \mathcal{L}_2,
\end{gather}
where $I_{c} \in \mathbb{R}^{1 \times H \times W}$ is the pseudo label generated by the Image Complexity Network (ICNet)~\cite{feng2022ic9600}.
$MSE$ denotes the Mean Square Error and $KL$ denotes the Kullback-Leibler divergence.
$\mathcal{S}$ and $\mathcal{LS}$ are $Softmax$ and $LogSoftmax$.
$\mathcal{M}$ denotes the flatten operation which reshapes the size of $I_{c}$ and $\hat{I_c}$ as $1 \times HW$.
$T$ is the temperature coefficient.
We use $\lambda_1$ and $\lambda_2$ as weights to combine the $MSE$ loss and $KL$ loss.
For the SS branch, we adopt the loss function $\mathcal{L}_{s}$ in PIDNet~\cite{xu2023pidnet}, and we finally get the overall object function as follows:
\begin{equation}
    \mathcal{L} = \gamma_1  \mathcal{L}_{c} + \gamma_2  \mathcal{L}_{s},
\end{equation}
where $\gamma_1$ and $\gamma_2$ are hyper-parameters to balance the segmentation loss and image complexity loss.

\begin{figure*}[ht]
    \centering
    \includegraphics*[width=1\textwidth]{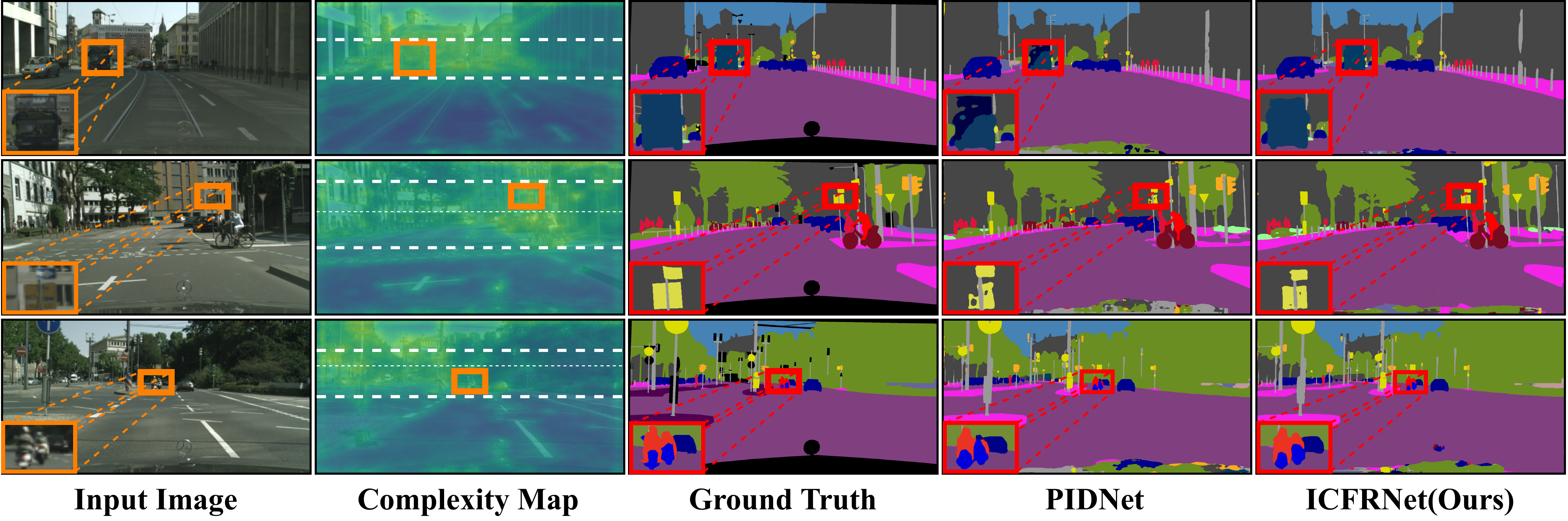}
    \caption{
    Visualization of semantic segmentation results on the Cityscapes val set.
    The maps in each row from left to right are the input image, predicted pixel-wise complexity map, ground truth segmentation results, results of PIDNet, and our results.
    Part of the high complexity regions are marked with orange boxes in the input images and the complexity maps.
    The corresponding regions in the ground truth and segmentation results are marked with red boxes and zoomed in for better illustration.}

    \label{fig:exp}
\end{figure*}

\begin{table}[t]\small
    \renewcommand\arraystretch{1.3}
    \centering
    \begin{tabular}{lccc}
        \toprule[1.5pt]
        Methods & GPU & \raggedright mIoU~($\%$)$\uparrow$  & FPS$\uparrow$ \\
        \midrule[1.2pt]
        \multicolumn{4}{c}{w/o Cityscapes pre-training} \\
        \hline
         BiSeNetV2\dag~\cite{yu2021bisenet} & GTX 1080Ti &  72.4 & 124.5\\
         BiSeNetV2-L\dag~\cite{yu2021bisenet} & GTX 1080Ti  & 73.2 & 32.7\\
         DDRNet-23-S~\cite{pan2022deep}& RTX 3090 &  74.7 & 182.4 \\
         PIDNet-S~\cite{xu2023pidnet}  & RTX 3090  &  75.3\textsuperscript{\tiny$\triangle$} & 127.9\textsuperscript{\tiny$\triangle$} \\
         \rowcolor{gray!20}\textbf{ICFRNet (Ours)}  & RTX 3090 & \textbf{75.9} & 114.2 \\
         \hline
         \multicolumn{4}{c}{w/ Cityscapes pre-training} \\
        \hline
         BiSeNetV2\dag \textsuperscript{*}~\cite{yu2021bisenet} & GTX 1080Ti  & 76.7 & 124.5\\
         BiSeNetV2-L\dag \textsuperscript{*}~\cite{yu2021bisenet} & GTX 1080Ti  & 78.5 & 32.7\\
         DDRNet-23-S\textsuperscript{*}~\cite{pan2022deep} & RTX 3090  &  78.6 & 182.4 \\
         PIDNet-S\textsuperscript{*}~\cite{xu2023pidnet}  & RTX 3090  &  79.4\textsuperscript{\tiny$\triangle$} & 127.9\textsuperscript{\tiny$\triangle$} \\
         \rowcolor{gray!20}\textbf{ICFRNet\textsuperscript{*} (Ours)}  & RTX 3090 & \textbf{80.0} & 114.2 \\
         \bottomrule[1.5pt]

    \end{tabular}
    \vspace{0.1cm}
    \caption{Comparison with state-of-the-art methods on the CamViD test set. 
    Methods marked with \textsuperscript{*} are pre-trained on the Cityscapes dataset. 
    \dag~means the speed of the method used TensorRT for acceleration. 
    \textsuperscript{\tiny$\triangle$} means our implementation.
    The resolution for comparison is $720 \times 960$.
    }
    \label{tab:camvid}
\end{table}

\section{Experimental Results}
\subsection{Settings}

\textbf{Datasets.} 
We perform experiments on two widely used datasets in real-time semantic segmentation tasks: Cityscapes~\cite{cordts2016cityscapes} and CamVid~\cite{BrostowSFC:ECCV08}. Both two datasets focus on the semantic segmentation on urban road scenes.
The Cityscapes dataset includes 19 annotated classes and consists of 5000 finely annotated images, with 2975, 500, and 1524 images allocated to the training, validation, and test sets. 
The resolution of images in Cityscapes is $\mathrm{1024 \times 2048}$.
The CamVid dataset includes 11 classes and 
comprises 701 images of resolution $\mathrm{720 \times 960}$, with 367, 101, and 233 images assigned to the training, validation, and test sets. 

\noindent\textbf{Implementation Details.} We use the weights of PIDNet pre-trained on the ImageNet~\cite{russakovsky2015imagenet} to initialize our model, and
use stochastic gradient descent (SGD) for optimization and apply the poly strategy with a power of 0.9 to schedule the learning rate.
Random crop, random scaling, and random horizontal flipping are adopted for data augmentation following~\cite{xu2023pidnet}.
We also use the warm-up strategy to accelerate the convergence in the early training stage and set the warm-up items to 1000 for both datasets.
The temperature coefficient $T$ is set to 0.6.
Parameters \{$\lambda_1$, $\lambda_2$, $\gamma_1$, $\gamma_2$\} are set to \{1, 10, 10, 1\} and \{1, 10, 1, 1\} for Cityscapes and CamViD datasets.
Batch size, initial learning rate, weight decay, training epochs, and crop size for Cityscapes and CamViD are configured as \{6, 0.005, 0.0005, 520, 1024$\times$1024\} and \{6, 0.005, 0.0007, 200, 720$\times$960\}.
Online hard example mining (Ohem) is also utilized following~\cite{xu2023pidnet}.
Our method is implemented in PyTorch 1.12 and tested on a single GPU of type NVIDIA RTX 3090.
%

\noindent\textbf{Evaluation metrics.} 
We evaluate our method in terms of accuracy and efficiency. 
For accuracy, we employ 
mean Intersection over Union (mIoU) to gauge our segmentation results. 
For efficiency, we incorporate FPS, GFLOPs, and Parameters as the metrics.
We adopt the same measurement protocol as PIDNet~\cite{xu2023pidnet} for FPS and the thop\footnote{https://github.com/Lyken17/pytorch-OpCounter.git} for the other two metrics. 

\subsection{Comparison With the State-of-the-Art Methods}
We first quantitatively compare our method with other SOTA methods on Cityscapes and CamViD datasets.

\noindent\textbf{Cityscapes.}
The comparison results are shown in Table \ref{tab:city}.
We report the results on both validation sets and test sets.
We test DDRNet-23-S and PIDNet-S on the Cityscapes server with their official validation model. While their testing models are trained with both training sets and validation sets, and our models are trained only with training sets.
The experiment results demonstrate that our ICFRNet improves the test accuracy by 0.6\% compared to PIDNet-S and 2.1\% compared to DDRNet-S while maintaining a competitive inference speed.
Note that we referred to partial results in~\cite{xu2023pidnet} for a comprehensive comparison.

\noindent\textbf{CamViD.}
Table \ref{tab:camvid} shows the comparison results on CamViD testing sets.
Since previous methods~\cite{pan2022deep,xu2023pidnet} employ models pre-trained on the Cityscapes for finetuning on CamVID, we report both results with and without the Cityscapes dataset pre-training.
For the finetuning with the Cityscapes, the initial learning rate is 0.001.
The experimental results demonstrate that without the Cityscapes dataset pre-training, our ICFRNet outperforms DDRNet-23-S and PIDNet-S by 1.2\% and 0.6\%.
With the Cityscapes dataset pre-training, ICFRNet outperforms DDRNet-23-S and PIDNet-S by 1.4\% and 0.6\%. 

The qualitative results are shown in Fig.~\ref{fig:exp}, 
We additionally zoom into regions marked with red boxes to provide a clearer illustration. It can be observed that our method excels in segmenting subtle and complex structures (e.g., the bus pixels in the first row of Fig.~\ref{fig:exp} are highly complex and misclassified by PIDNet, but are well segmented by our ICFRNet). This observation indicates that our method benefits from the exploration of image complexity information, effectively addressing those challenging regions.

\begin{table}[t]\small
    \renewcommand\arraystretch{1.3}
    \centering
    \begin{tabular}{c|cccc|p{1.2cm}<{\centering}p{1.0cm}<{\centering}}
        \toprule[1.5pt]
        CP & ~\cite{feng2022ic9600} & Concat  & ICGA & IC &mIoU(\%)~$\uparrow$ & FPS~$\uparrow$  \\
        \midrule[1.2pt]
        \multirow{6}{*}{w/o} & \Checkmark  &         &           &               &  74.9    &  127.9 \\
        \cline{2-7}
         & & \Checkmark &                  &               &  75.2    &  117.5 \\
        &  \cellcolor{gray!20} & \cellcolor{gray!20}\Checkmark &     \cellcolor{gray!20}    &  \cellcolor{gray!20} \Checkmark    & \cellcolor{gray!20} 75.0     & \cellcolor{gray!20} 117.5 \\
             & &  & \Checkmark    &               & 75.5      & 114.2  \\
          &  \cellcolor{gray!20}  &  \cellcolor{gray!20}  & \cellcolor{gray!20} \Checkmark    & \cellcolor{gray!20} \Checkmark     & \cellcolor{gray!20} 75.9      & \cellcolor{gray!20} 114.2  \\
        \hline
         w  &  &      & \Checkmark    &\Checkmark     & \textbf{80.0}      & 114.2  \\
        
        \bottomrule[1.5pt]
   
    \end{tabular}
    \vspace{0.1cm}
    \caption{Ablation study on CamViD dataset.
    CP denotes Cityscapes pre-training.
    Method in~\cite{feng2022ic9600} directly multiplies the image complexity map produced by ICNet~\cite{feng2022ic9600} with the pixel-wise cross-entropy loss. 
    Concat denotes that the segmentation features are concatenated with the image complexity feature.
    IC denotes the model has IC Loss as supervision.}
    \label{tab:ablation}
\end{table}

\subsection{Ablation Studies}
We carry out ablation studies to validate the effectiveness of framework design, reported in Table \ref{tab:ablation}.
We first apply the method used in~\cite{feng2022ic9600} to the base segmentation model ($i.e.$, SS Branch), which directly multiplies the image complexity map with the pixel-wise cross entropy loss, and presents a lower segmentation accuracy.
We further replace the ICGA module with concatenation operation under conditions of whether IC Loss exists.
The concatenation operation with IC Loss is 0.2\% lower than the results without that, while ICGA with IC Loss gains a 0.4\% promotion.
This indicates that simply concatenating image complexity features will instead interfere with the segmentation, while our ICGA module can effectively utilize the complexity information and gain better performance.
On the other hand, compared to concatenation, ICGA achieves a 0.3\% improvement in mIoU under conditions without IC Loss and a 0.9\% improvement with IC Loss.
This further demonstrates that our ICGA module can handle conditions both with and without image complexity information, and enables better performance with image complexity guidance, meanwhile maintaining a competitive inference speed.
Besides, with the Cityscapes pre-training, segmentation quality can be further improved on the CamViD dataset.

\section{Conclusion}
We establish a dual-task framework named ICFRNet and devise the Image Complexity Guided Attention (ICGA) module to incorporate image complexity features as spatial attention guidance for refining segmentation features. The network is optimized in terms of both segmentation and image complexity prediction tasks with a combined loss function. Experimental results show that our proposed framework with the ICGA module can effectively improve segmentation accuracy while maintaining a competitive inference speed.



\bibliographystyle{IEEEtran}
\bibliography{reference}

\end{document}